\documentclass[letterpaper]{article} 
\usepackage{aaai2027}  
\usepackage[hyphens]{url}  
\usepackage{graphicx} 
\usepackage{natbib}  
\usepackage{caption} 
\usepackage{algorithm}
\usepackage{algorithmic}

\usepackage{newfloat}
\usepackage{listings}
\DeclareCaptionStyle{ruled}{labelfont=normalfont,labelsep=colon,strut=off} 
\floatstyle{ruled}
\newfloat{listing}{tb}{lst}{}
\floatname{listing}{Listing}

\usepackage{booktabs}
\usepackage{multirow}
\usepackage{amssymb}
\usepackage{pifont}

\makeatletter
\newcommand\aaai@dag{{\usefont{OMS}{cmsy}{m}{n}\symbol{121}}}   
\newcommand\aaai@ddag{{\usefont{OMS}{cmsy}{m}{n}\symbol{122}}}  
\renewcommand*{\@fnsymbol}[1]{%
  \ifcase#1\or *\or \aaai@dag\or \aaai@ddag\or \textsection\or
  \textparagraph\or \textbardbl\or **\or \aaai@dag\aaai@dag\or
  \aaai@ddag\aaai@ddag\else\@ctrerr\fi}%
\makeatother
\usepackage{xspace}
\providecommand{\sys}{\textsc{MSEval}\xspace}

\newcommand{\marksizescale}{1.18}
\newcommand{\cmark}{\scalebox{\marksizescale}{\ding{51}}}
\newcommand{\hmark}{%
  \kern-0.18em%
  \scalebox{\marksizescale}{%
    \ding{51}%
    \kern-0.45em%
    \raisebox{1.3ex}{%
      \rotatebox{-58}{%
        \rule{0.95ex}{0.75pt}%
      }%
    }%
  }%
}
\newcommand{\xmark}{\scalebox{\marksizescale}{\ding{55}}}

\usepackage{tikz}
\usetikzlibrary{positioning,arrows.meta,fit,backgrounds,calc}

\title{An Empirical Study of Coordination Mode as the First-Class Citizen in From-Scratch Multi-Agent Coding}
\author {
    Yanyu Ren\textsuperscript{\rm 1}\equalcontrib,
    Yunfeng Bai\textsuperscript{\rm 1}\equalcontrib,
    Xizheng Wang\textsuperscript{\rm 2}\corresponding,
    Li Chen\textsuperscript{\rm 2}\corresponding,
    Dan Li\textsuperscript{\rm 1}
}
\affiliations {
    \textsuperscript{\rm 1}Tsinghua University \hspace{3.5em} \textsuperscript{\rm 2}Zhongguancun Laboratory
}

\nocopyright

\begin{document}

\maketitle
\begin{abstract}
Multi-agent vibe coding promises to accelerate software development, yet existing benchmarks rely on synthetic environments that ignore practical time and monetary costs, conflate reasoning with communication, and reward only superficial completion. We introduce multi-agent from-scratch evaluation benchmark, \sys, evaluating multi-agent coding on real-world tasks. Grounded in 10 authentic, full-stack projects across 10 domains, \sys scores performance using hierarchical requirements and deterministic rubrics.

Its execution engine, \textsc{LegoGent}, tests 10 collaboration topologies where agents coordinate via periodic \emph{sync} intervals and deploy through native CI/CD pipelines. Concurrently, the automated grader \textsc{TAgent} dynamically probes implementations to jointly measure functional success, latency, and prefix-cached token cost. Across 100 runs, \sys reveals that organizational topology rivals model capability in shaping the speed--cost--quality trade-off. For identical tasks and models, varying the topology shifts scores by over 30 points and doubles wall-clock time. Structured pipelines converge fastest with the highest quality, whereas heavy managerial oversight degrades performance. Ultimately, \sys establishes a rigorous, reproducible standard for measuring how multi-agent teams actually build software. The benchmark is released at  \url{https://github.com/robinren03/MSEval}.
\end{abstract}

\section{Introduction}
Frontier coding agents have made repository-level programming more capable, shifting the field from toy function synthesis toward long-horizon, execution-grounded software work \cite{swebench,deepswe,frontiercode}. Yet most evaluations still ask whether one agent produces an acceptable patch. They do not measure what increasingly matters in practice: how a team of agents divides labor, builds a project from scratch without pre-defined judges, spends time and tokens, and improves after concrete feedback.

Multi-agent systems promise faster software delivery by assigning specialized workers to parallel subtasks, but this promise is fragile. Poorly bounded communication can amplify semantic drift; heavy orchestration can become a bottleneck; and self-reported completion is often detached from runnable behavior. Existing multi-agent benchmarks expose some communication patterns \cite{multiagentbench}, but they are often simulated, detached from continuous integration and continuous delivery (CI/CD) deployment, and too coarse to explain why one collaboration mode succeeds while another fails on the same project and model.

We introduce \sys, a benchmark for quantitatively evaluating multi-agent vibe coding on real-world software tasks. \sys is grounded in a university software-engineering setting where groups of three to four developers iteratively build full-stack projects and receive teaching-assistant (TA) feedback. It contains 10 rubric-scored web projects, 10 collaboration topologies implemented in \textsc{LegoGent}, and an automated evaluator, \textsc{TAgent}, that discovers and probes each deployed artifact before returning weighted feedback for the next refinement round.

\sys is designed around four properties that are missing in combination from prior coding-agent evaluations:
\begin{itemize}
    \item \textbf{Diverse Collaboration.} The same project can be run under feature squads, layer specialists, pipeline handoff, swarming, rotation, project-manager (PM) oversight, quality-assurance~(QA)-first, Push Request~(PR)-style review, adversarial testing, and competitive teams.
    \item \textbf{Deployment-Grounded Evaluation.} \sys builds and serves artifacts through independent endpoints, and graded through live user interface (UI), application programming interface (API), and code checks rather than final-text inspection.
    \item \textbf{Iterative Feedback.} Agents receive \textsc{TAgent}'s automatically-generated scores and comments for up to three refinement rounds, exposing convergence and regression instead of only first-attempt success.
    \item \textbf{First-Class Metrics.} Wall-clock latency and prefix-cached token usage are logged with the functional score, making coordination overhead measurable.
\end{itemize}

Our key contributions are:
\begin{itemize}
    \item a 100-case benchmark grid over 10 realistic projects and 10 collaboration modes, with project specifications, collaboration-mode definitions, source code, metrics, and analysis scripts provided in the supplementary materials;
    \item \textsc{LegoGent}, an activation-driven multi-agent runtime with bounded synchronization, explicit peer messaging, CI/CD deployment, and behavioral completion checks;
    \item \textsc{TAgent}, a document-driven grader that discovers UI/API/code surfaces in unseen implementations and returns weighted, actionable feedback;
    \item an empirical study showing that topology, model capability, and project domain jointly determine the speed--cost--quality trade-off.
\end{itemize}

\begin{figure*}[t]
    \centering
    \includegraphics[width=0.83\linewidth]{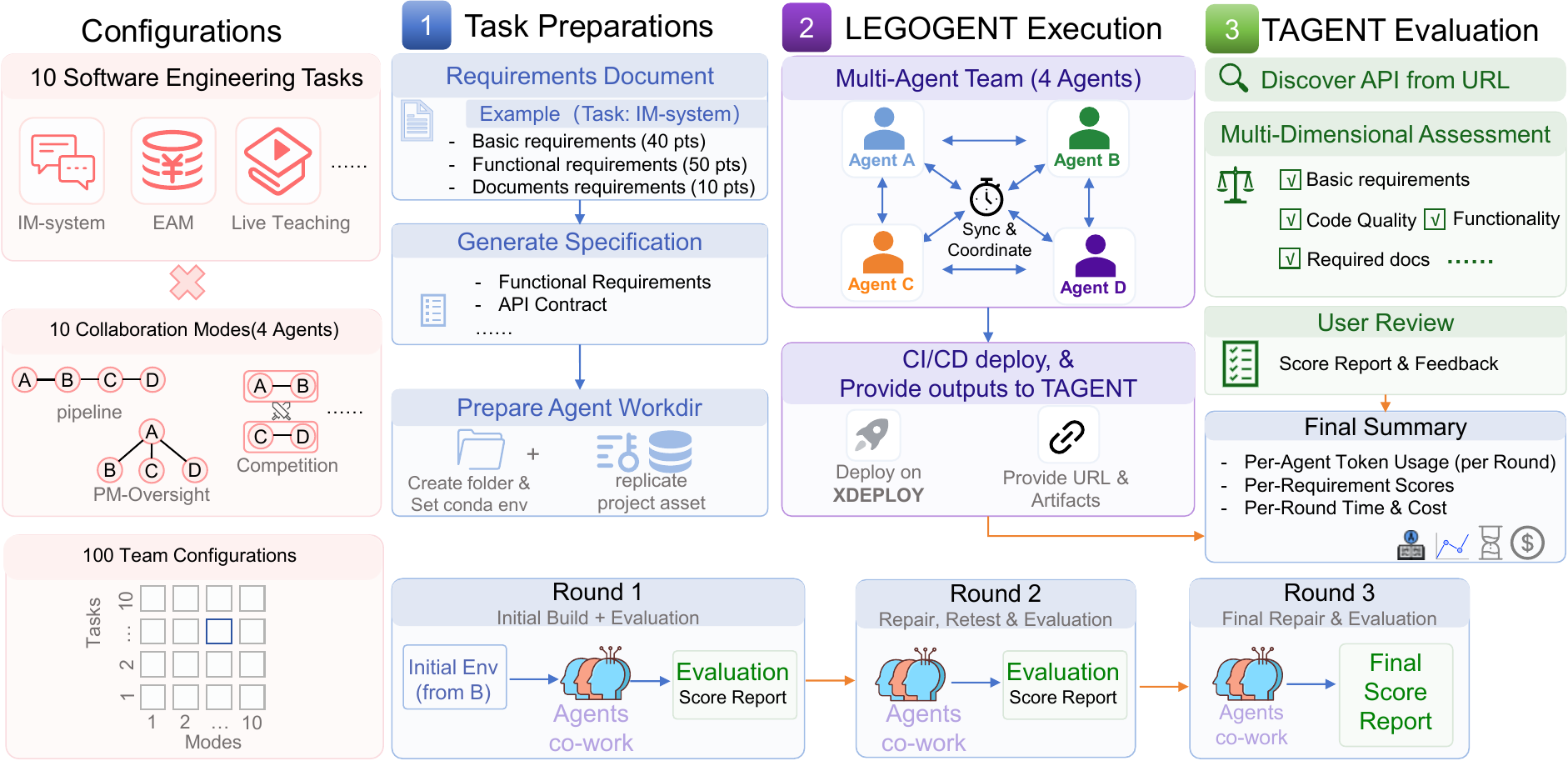}
    \caption{\sys overview. \sys builds and evaluates a web app from requirement document without detailed API specifications and evaluate the projects comprehensively.}
    \label{fig:architecture}
\end{figure*}

\section{Background and Related Work}

\subsection{Coding Agents}
Coding-agent evaluation has moved from local synthesis to repository-scale software work. HumanEval and MBPP measure function-level generation, while RepoBench and SWE-bench move toward multi-file context and real GitHub issues \cite{humaneval,mbpp,repobench,swebench}. Agent systems such as SWE-agent, Agentless, AutoCodeRover, and OpenHands further show that language models can navigate repositories, run tools, and repair defects \cite{sweagent,agentless,autocoderover,openhands}. DeepSWE and FrontierCode push this line toward longer-horizon engineering tasks \cite{deepswe,frontiercode}. This progress is important, but it leaves a different question unanswered. Most coding benchmarks still evaluate a single agent's repair behavior, patch validity, or code-quality outcome. They do not ask whether a team of agents can start from requirements, divide work, deploy a full-stack system, and improve after concrete feedback while spending time and tokens. That is the gap \sys targets: the unit of measurement is no longer only a model's coding skill, but the whole software-delivery loop around the model.

\subsection{Multi-Agent Benchmarks}
Existing multi-agent benchmarks also leave this gap open, but for a different reason. AgentsNet and MultiAgentBench study collaboration and competition across communication structures, yet their tasks are more abstract than deployed software delivery \cite{agentsnet,multiagentbench}. They can expose communication behavior, but not whether a collaboration policy produces a runnable application, respects interface boundaries, or survives integration. CloudDevBench evaluates multi-agent repository development from specifications using CloudMAS, a multi-agent coding assistant, over 50 test cases~\cite{clouddevbench}. However, CloudDevBench uses a fixed collaboration mode and takes advantage of detailed task specifications with pre-defined APIs. It is therefore only partially from scratch: a fine-grained design protocol should be given for each project, rather than a coarse-grained requirement document. TheAgentCompany broadens the task mix, but it does not control collaboration policy inside a real software-delivery loop \cite{theagentcompany}. These designs cannot separate weak coding ability from duplicated work, stale shared context, bad handoffs, deployment friction, or inefficient evaluation.

\subsection{Motivation for \sys}
\sys is designed around the missing combination: realistic from-scratch projects, multiple collaboration modes, deployment-grounded scoring, and cost-aware feedback in one controlled setup. With an end-to-end artifact typically with 2000+ lines of Python code to be generated, \sys is not sensitive to data contamination of LLM. The benchmark keeps the project, model, deployment path, and rubric fixed while varying the collaboration topology. This makes topology measurable rather than incidental: a low score can be interpreted together with the run's time, token use, deployment behavior, and feedback trajectory. \sys is a benchmark for multi-agent software work rather than a leaderboard for one final artifact. Table~\ref{tab:benchmark_comparison} summarizes \sys's features.

\begin{table}[!t]
\centering
\small
\setlength{\tabcolsep}{1mm}
\renewcommand{\arraystretch}{0.96}
\begin{tabular*}{\columnwidth}{@{\extracolsep{\fill}}p{0.25\columnwidth} c c c c c@{}}
\toprule
\textbf{Work} & \textbf{Code} & \shortstack{\textbf{From}\\\textbf{Scratch}} & \shortstack{\textbf{Multi}\\\textbf{Agent}} & \shortstack{\textbf{Collab.}\\\textbf{Focus}} & \shortstack{\textbf{Cost}\\\textbf{Aware}} \\
\midrule
SWE-bench & \cmark & \xmark & \xmark & \xmark & \xmark \\
MultiAgent\allowbreak Bench & \xmark & \xmark & \cmark & \cmark & \hmark \\
CloudDev\allowbreak Bench & \cmark & \hmark & \cmark & \xmark & \xmark \\
Deep\allowbreak SWE & \cmark & \xmark & \xmark & \xmark & \xmark \\
Frontier\allowbreak Code & \cmark & \xmark & \xmark & \xmark & \xmark \\
TheAgent\allowbreak Company & \hmark & \xmark & \cmark & \xmark & \xmark \\
\textbf{\sys (ours)} & \textbf{\cmark} & \textbf{\cmark} & \textbf{\cmark} & \textbf{\cmark} & \textbf{\cmark} \\
\bottomrule
\end{tabular*}
\caption{Comparison of \sys against existing benchmarks. \sys is the first benchmark that examine the impact of multi-agent coordination on from-scratch coding.}
\label{tab:benchmark_comparison}
\end{table}

\section{Methodology}

\subsection{Benchmark Framework Overview}
The \sys benchmark is a three-stage pipeline for measuring how multi-agent teams build deployable software under different coordination policies.

\noindent\textbf{Data foundation.} We curate 10 full-scale software-engineering projects from a corpus of university capstone assignments developed since 2018. The projects cover real-time messaging, enterprise asset management, crowdsourcing, requirement tracking, image processing, e-commerce, creator analytics, role-based access control (RBAC), news search, and online live teaching during pandemic. Each requirement document decomposes into six to eight modules and 30--45 weighted line items, normalized to a deterministic 100-point rubric.

\noindent\textbf{Pipeline execution.} \textsc{LegoGent} instantiates a selected collaboration mode, launches agents with role-specific prompts, monitors progress, and sends the generated repository to \textsc{XDeploy}, which builds and stages the application through a native CI/CD path with GitLab and SonarQube~\cite{sonarqube}. This makes deployment failure, not just code incompleteness, part of the measured outcome.
\noindent\textbf{Metrics.} \textsc{TAgent} grades the staged artifact against the rubric and produces a completion score on the 0--100 scale. \textsc{LegoGent} records cumulative wall-clock time and token usage. \sys joins quality, speed, and cost for every round.

\noindent\textbf{The collaboration mode is the only independent variable.} The benchmark treats a collaboration mode as the independent variable rather than a prompt flourish. For a fixed project and model family, every run receives the same requirement document, round budget, deployment path, and \textsc{TAgent} rubric. What changes is the organizational policy: who owns each artifact, which agents may run concurrently, where handoffs occur, who can reject a completion claim, and how feedback is routed into the next round. This design lets the evaluation distinguish a weak model from a weak topology. A low score after a failed deploy is not collapsed into a generic failure; it is paired with the time and token trace that shows whether the team spent the round coding, coordinating, waiting for a slow model, or repairing a bad merge.

\subsection{Multi-Agent Execution via \textsc{LegoGent}}

\begin{figure}[t]
    \centering
    \includegraphics[width=\columnwidth]{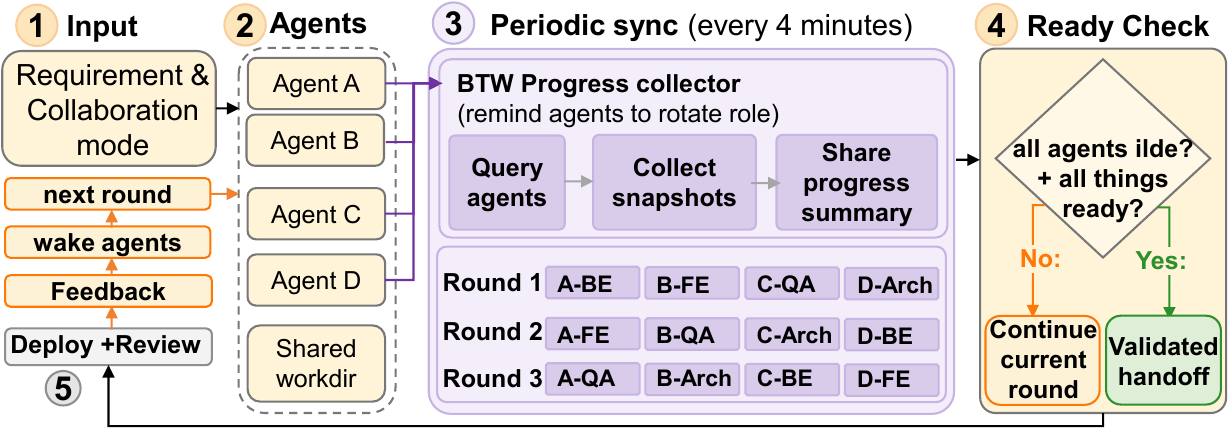}
    \caption{An illustration of \textsc{LegoGent} execution under the rotation collaboration mode, where the agents switch their roles in each round. BE and FE mean backend and frontend.}
    \label{fig:legogent}
\end{figure}

\textsc{LegoGent} is the runtime that turns a collaboration mode into an executable team. Each agent runs as an isolated, resumable process with file and shell tools in a shared workspace, which framework-side validators guard by checking ports, HTTP behavior, and artifacts before accepting completion claims. Rather than inferring a fixed dependency graph, \textsc{LegoGent} runs a periodic sync loop (Figure~\ref{fig:legogent}): agents work concurrently while a progress collector polls them and broadcasts a shared snapshot. A completion gate then checks whether all agents are idle and the required artifacts are ready; if so, the handoff is deployed and reviewed and its feedback opens the next round, otherwise the round continues.

On this runtime we implement 10 collaboration modes. Some modes are handoff-heavy, such as pipeline and PM oversight pass work through fixed stages, swarming claims tasks dynamically; while the other modes often encounter merge conflicts, open-source review, adversarial testing, and competitive dual teams which frequently face disagreement. The supplementary material details each mode.

Coordination uses two bounded channels. A passive \emph{sync} round runs every four minutes: each agent reports completed work, test evidence, blockers, and next steps, and the runtime broadcasts a consolidated state snapshot to the team. An active mailbox supports targeted peer questions and handoffs between sync rounds. This design gives parallel workers a shared view without collapsing them into one monolithic context, and it makes communication overhead observable through time and token logs.

\noindent\textbf{Mode templates make parallel work comparable.} A mode template specifies four concrete objects: an ownership map, an activation schedule, a decision rule, and a required handoff artifact. Feature squads parallelize by user-facing module; layer specialists parallelize within frontend and backend layers and synchronize at API contracts; swarming allows dynamic task claiming; competitive mode runs redundant implementations and selects a winner. Pipeline mode is intentionally almost serial, so it acts as a coordination-control condition, trading parallel speed for a clean artifact boundary between architect, backend, frontend, and QA. As all modes write into the same repository and pass the same deployment validator, the benchmark can observe whether parallelism actually reduces wall-clock time or merely creates duplicate context and merge repair.

\noindent\textbf{\textsc{LegoGent} enables validated artifacts as unit of collaboration.} The useful unit of collaboration is validated artifacts, such as a schema, route contract, integration point, test report, or deployable repository state. \textsc{LegoGent} crops the overall failure report, and decides which agent is responsible for the errors before any code changes. On the other hand, \sys also exposes the failure of certain modes with diffuse ownership, where the agents can post a competitive final score yet be unattractive as an agentic workflow, since its score is bought with extra token or time cost.

\subsection{Automated Evaluation via \textsc{TAgent}}

\begin{figure}[t]
    \centering
    \includegraphics[width=\columnwidth]{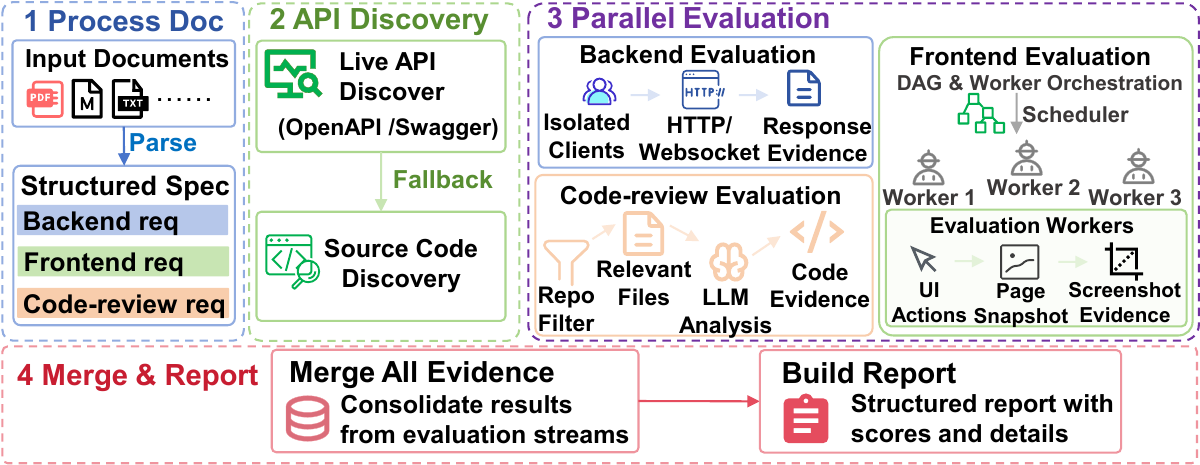}
    \caption{\textsc{TAgent} parses a requirement document into weighted checks, discovers the implementation, and merges parallel UI, API, and code evidence into a score.}
    \label{fig:tutorgent}
\end{figure}

\textsc{TAgent} acts as an automated TA for heterogeneous agent-generated projects. It first parses each raw requirement document into a structured YAML rubric with weighted scenarios and pass/partial/fail criteria. During evaluation it receives the staged URL and repository, then discovers the implementation surface before probing it. Requirements may be routed to three evaluator classes:
\begin{itemize}
    \item \textbf{UI:} Playwright drives the live site, reads the accessibility tree or DOM summary, chooses observed selectors, and judges captured states against the rubric.
    \item \textbf{API:} the grader probes OpenAPI/Swagger when available, otherwise reconstructs routes from source files across common frameworks, then issues authenticated HTTP sequences.
    \item \textbf{Code:} directory filtering, targeted search, and LLM inspection verify constraints that are not externally visible, such as password hashing or CI/CD configuration.
\end{itemize}

\noindent\textbf{Parallel UI scheduling.} API and code checks are mostly stateless and fan out under a concurrency cap. UI checks are stateful: testing a group chat may require registration, login, friend creation, and room entry. Running every check independently repeats this setup, while naive parallelism duplicates it across workers. \textsc{TAgent} therefore asks an LLM to declare only blocking dependencies among UI requirements, forms a task DAG, and partitions it across browser workers with an integer-programming objective that minimizes makespan plus duplicate setup. Prerequisites are co-located with dependents, shared setup is built once per worker, unreachable prerequisites cascade-skip their dependents, and duplicate probes are merged by taking the best verdict. This is the evaluator-side analogue of efficient multi-agent work: useful parallelism comes from eliminating redundant setup, not merely launching more workers.

\noindent\textbf{Feedback as repair guidance.} A final scalar is insufficient for multi-round coding because it does not tell agents where the dependency chain broke. \textsc{TAgent} therefore emits item-level evidence: the observed UI state, API response, code location, failed prerequisite, and the rubric weight affected. \textsc{LegoGent} passes this back as a ranked repair agenda, prioritizing high-weight root causes over low-weight symptoms---fixing registration can unlock downstream friend, chat, and group checks, while a cosmetic chat bubble cannot compensate for a missing message API. This is why round-level improvement is interpretable rather than accidental.

\noindent\textbf{Evaluator parallelism is deliberately conservative.} Because web applications are not pure functions, \textsc{TAgent} parallelizes probes only after discovering dependencies: a worker that deletes a friend, submits a group application, or changes profile data can poison another check sharing the same account state. The scheduler thus prefers independent browser sessions for destructive flows and shared sessions for long setup chains, trading wall-clock time for verdicts that do not depend on a lucky execution order. The same principle motivates the benchmark itself: parallel agents are useful when their side effects are isolated, and risky when they touch shared state without a merge protocol.

Scores are aggregated by rubric weight to a 100-point Functional Completion Score. Together with \textsc{LegoGent}'s latency and token logs, this report becomes the feedback signal for the next refinement round.

\section{Evaluation}

We evaluate the models in \sys on four Intel(R) Xeon(R) Gold 6430 processors. Each run allows up to three refinement rounds; a round builds or patches the project, deploys it, runs \textsc{TAgent}, and returns scored feedback. Deployment, testing, and evaluation use isolated port ranges so concurrent tests do not interfere with one another. We report Functional Completion Score (0--100), best-round wall-clock minutes, USD cost, and raw token usage at the best observed round. Cost is computed from the provider price tables over new-input, cache-read, cache-create, and output token fields listed in the supplementary material~\cite{anthropic_opus48_pricing,openai_gpt55_pricing,deepseek_v4_pricing,zai_pricing,qwen_pricing}.

\subsection{Results by Model}

Table~\ref{tab:eval_models} reports the instant-messaging project across all ten modes and five models. Each block gives a compact R1/R2/R3 score cell with best-round minutes, USD cost, and million tokens, so the table reads as a quality--latency--budget frontier rather than a final-score leaderboard. Claude Opus 4.8~\cite{anthropic_claude48} tops this project, GPT-5.5~\cite{openai_gpt55} stays close at much lower budget, DeepSeek v4 Pro~\cite{deepseek_v4} is the stronger full-grid baseline, and GLM-5.2~\cite{glm52} reaches the highest full-grid average score but is the slowest by a wide margin. Qwen3.6-Flash~\cite{qwen36flash} stays at 0 across all ten modes (final round 39--270 min, \$9.04--\$61.25), so we omit it from the results.

\begin{table*}[!t]
\centering
\small
\setlength{\tabcolsep}{1mm}
\renewcommand{\arraystretch}{0.9}
\begin{tabular*}{\textwidth}{@{\extracolsep{\fill}}l*{20}{c}@{}}
\toprule
\multirow{2}{*}{\textbf{Mode}} & \multicolumn{4}{c}{\textbf{Claude Opus 4.8}} & \multicolumn{4}{c}{\textbf{GPT-5.5}} & \multicolumn{4}{c}{\textbf{DeepSeek v4 Flash}} & \multicolumn{4}{c}{\textbf{DeepSeek v4 Pro}} & \multicolumn{4}{c}{\textbf{GLM-5.2}} \\
 & \textbf{Score} & \textbf{T} & \textbf{C} & \textbf{Tok} & \textbf{Score} & \textbf{T} & \textbf{C} & \textbf{Tok} & \textbf{Score} & \textbf{T} & \textbf{C} & \textbf{Tok} & \textbf{Score} & \textbf{T} & \textbf{C} & \textbf{Tok} & \textbf{Score} & \textbf{T} & \textbf{C} & \textbf{Tok} \\
\cmidrule(lr){2-5} \cmidrule(lr){6-9} \cmidrule(lr){10-13} \cmidrule(lr){14-17} \cmidrule(l){18-21}
Feat. & 80/94/\textbf{97} & \textbf{110} & \textbf{654} & \textbf{114} & 81/\textbf{91}/79 & 86 & 153 & 31 & 0/54/\textbf{68} & 101 & 3 & 474 & 52/57/\textbf{83} & 78 & 8 & 572 & 40/\textbf{73}/61 & 53 & 42 & 123 \\
Layer & 74/73/\textbf{85} & 102 & 191 & 169 & 79/93/\textbf{94} & \textbf{82} & \textbf{139} & \textbf{28} & 63/0/\textbf{70} & 108 & 4 & 602 & 28/68/\textbf{87} & 101 & 6 & 359 & 51/71/\textbf{72} & 129 & 75 & 222 \\
Pipe. & 65/79/\textbf{91} & 143 & 214 & 172 & 72/83/\textbf{92} & \textbf{94} & \textbf{159} & \textbf{32} & 0/32/\textbf{83} & 67 & 3 & 348 & 41/80/\textbf{90} & 111 & 9 & 723 & 72/0/\textbf{78} & 144 & 84 & 246 \\
PM & 72/91/\textbf{94} & \textbf{176} & \textbf{834} & \textbf{157} & 79/84/\textbf{86} & 119 & 274 & 54 & 54/\textbf{89}/73 & 94 & 3 & 492 & 78/\textbf{81}/66 & 99 & 6 & 340 & 53/72/\textbf{82} & 142 & 148 & 406 \\
QA & 5/13/\textbf{72} & 102 & 118 & 140 & 83/81/\textbf{92} & \textbf{99} & \textbf{247} & \textbf{49} & 71/74/\textbf{83} & 108 & 4 & 466 & 70/74/\textbf{89} & 85 & 8 & 583 & 59/\textbf{64}/54 & 94 & 63 & 174 \\
Swarm & 76/84/\textbf{85} & 151 & 988 & 226 & 84/86/\textbf{87} & 78 & 255 & 51 & 0/64/\textbf{72} & 121 & 4 & 524 & 46/72/\textbf{88} & \textbf{103} & \textbf{10} & \textbf{669} & 0/82/\textbf{87} & 111 & 110 & 298 \\
Rot. & 76/82/\textbf{84} & 143 & 872 & 193 & 62/\textbf{93}/83 & \textbf{86} & \textbf{295} & \textbf{59} & 45/38/\textbf{47} & 105 & 4 & 445 & 67/76/\textbf{85} & 107 & 10 & 653 & 39/69/\textbf{79} & 154 & 60 & 169 \\
OSS & 0/75/\textbf{87} & 282 & 190 & 33 & 56/85/\textbf{90} & \textbf{278} & \textbf{302} & \textbf{60} & 47/\textbf{73}/69 & 62 & 3 & 429 & 0/27/\textbf{43} & 126 & 11 & 744 & 0/\textbf{70}/67 & 180 & 93 & 241 \\
Adv. & 63/79/\textbf{86} & 192 & 204 & 34 & 62/\textbf{96}/93 & \textbf{74} & \textbf{138} & \textbf{28} & 36/68/\textbf{82} & 106 & 5 & 664 & \textbf{70}/65/70 & 101 & 9 & 390 & 60/76/\textbf{92} & 165 & 101 & 277 \\
Comp. & 65/76/\textbf{77} & 180 & 196 & 33 & 0/75/\textbf{89} & \textbf{172} & \textbf{256} & \textbf{51} & 0/77/\textbf{86} & 109 & 5 & 649 & 68/74/\textbf{77} & 133 & 11 & 679 & 0/71/\textbf{86} & 174 & 95 & 259 \\
\bottomrule
\end{tabular*}
\caption{Metrics of instant-messaging project. We display the R1/R2/R3 scores with the coding time (in minutes), cost (in USD), and total input and output tokens (in million); bold marks the row-best model and its associated best-round metrics.}
\label{tab:eval_models}
\end{table*}

\begin{figure*}[!t]
\centering
\includegraphics[width=0.88\textwidth]{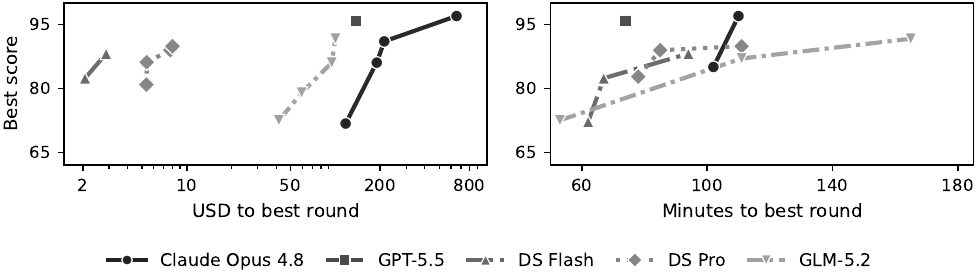}
\caption{Pareto-frontier collaboration modes on instant messaging, scores against USD cost and wall-clock time.}
\label{fig:model_cost_frontier}
\end{figure*}

Figure~\ref{fig:model_cost_frontier} makes the comparison cost-aware without collapsing each model to a single dot or drawing dominated modes. Claude Opus 4.8 reaches the highest score (97) but needs 110 minutes and about \$654, while GPT-5.5 is close (96) at 74 minutes and \$138; among full-grid models GLM-5.2 slightly edges DeepSeek v4 Pro but at roughly 2.6x the wall-clock time and 8x the cost. The practical question is therefore not only which model is best, but which frontier mode buys score cheaply and when the controller should stop spending.

\subsection{Results by Collaboration Mode}

Table~\ref{tab:eval_projects} tests whether a mode transfers beyond instant messaging, reporting the 10-project average per mode and model. The compact R1/R2/R3 column keeps the refinement trajectory visible, while T, C, and Tok show whether the best round is reached quickly, cheaply, or through heavier token use. A mode can look strong on one project; the aggregate tests whether it survives diverse projects with different inter-module interaction patterns.

\begin{table}[!t]
\centering
\small
\setlength{\tabcolsep}{3mm}
\renewcommand{\arraystretch}{0.86}
\begin{tabular}{@{}lcccc@{}}
\toprule
\textbf{Mode} & \textbf{Score} & \textbf{T} & \textbf{C} & \textbf{Tok} \\
\midrule
\multicolumn{5}{l}{\textbf{DeepSeek v4 Flash}} \\
\midrule
Feature squad & 24.9/50.3/71.3 & 80 & 3 & 441 \\
Layer specialists & 21.2/44.8/67.6 & 100 & 3 & 422 \\
Pipeline & 34.6/58.6/71.3 & 78 & 2 & 362 \\
PM oversight & 30.2/58.9/75.2 & 99 & 3 & 444 \\
QA-first & 52.0/70.2/\textbf{80.6} & \textbf{96} & \textbf{3} & \textbf{460} \\
Swarming & 25.2/44.8/62.9 & 84 & 3 & 512 \\
Rotation & 36.6/68.3/73.8 & 80 & 3 & 457 \\
Open-source & 21.7/49.5/72.0 & 77 & 3 & 429 \\
Adversarial & 28.2/44.9/67.7 & 90 & 3 & 460 \\
Competitive & 40.7/70.3/80.0 & 108 & 4 & 566 \\
\midrule
\multicolumn{5}{l}{\textbf{DeepSeek v4 Pro}} \\
\midrule
Feature squad & 37.9/67.6/83.9 & 83 & 8 & 537 \\
Layer specialists & 42.5/68.5/81.0 & 92 & 8 & 466 \\
Pipeline & 59.5/80.1/83.7 & 111 & 7 & 491 \\
PM oversight & 43.7/74.5/78.3 & 96 & 7 & 423 \\
QA-first & 52.0/75.9/85.5 & 95 & 8 & 514 \\
Swarming & 63.8/80.6/83.0 & 76 & 8 & 554 \\
Rotation & 51.6/73.2/\textbf{86.6} & \textbf{99} & \textbf{9} & \textbf{553} \\
Open-source & 38.5/58.4/69.2 & 92 & 9 & 622 \\
Adversarial & 42.0/68.2/70.2 & 89 & 8 & 484 \\
Competitive & 46.5/63.1/78.2 & 105 & 8 & 518 \\
\midrule
\multicolumn{5}{l}{\textbf{GLM-5.2}} \\
\midrule
Feature squad & 37.1/66.2/85.4 & 231 & 45 & 128 \\
Layer specialists & 41.2/61.1/77.2 & 254 & 49 & 140 \\
Pipeline & 61.1/76.0/\textbf{87.6} & \textbf{246} & \textbf{57} & \textbf{162} \\
PM oversight & 30.5/65.5/85.3 & 249 & 72 & 205 \\
QA-first & 57.4/68.9/80.7 & 235 & 56 & 157 \\
Swarming & 35.4/77.6/86.4 & 250 & 61 & 172 \\
Rotation & 56.0/72.7/86.9 & 257 & 75 & 188 \\
Open-source & 8.0/37.5/69.9 & 245 & 60 & 166 \\
Adversarial & 38.8/75.8/82.6 & 235 & 95 & 193 \\
Competitive & 8.1/61.3/85.8 & 263 & 67 & 188 \\
\bottomrule
\end{tabular}
\caption{Average metrics over all 10 projects. The metric definition and bold convention are the same as Table~\ref{tab:eval_models}.}
\label{tab:eval_projects}
\end{table}

\begin{figure}[!t]
\centering
\includegraphics[width=\columnwidth]{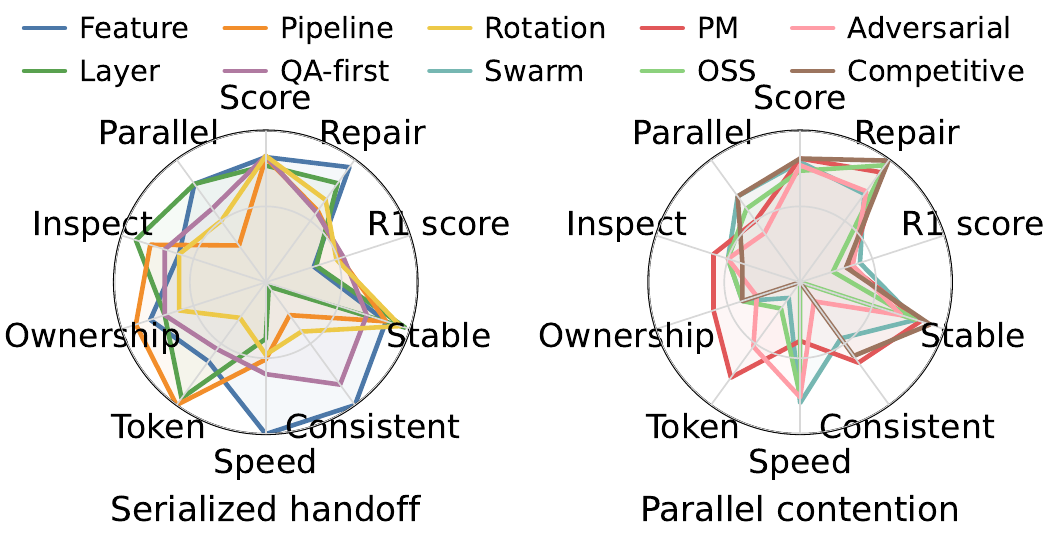}
\caption{Collaboration-mode radar on ten normalized axes. Higher values indicate stronger capability or lower cost.}
\label{fig:mode_radar}
\end{figure}

We also profile each mode as an execution signature rather than a score-only outcome in Figure~\ref{fig:mode_radar}. All ten axes are normalized to 0--10 (higher is better): seven from the logs (best/R1 score, R1-to-R3 gain, R2-to-R3 stability, cross-run consistency, inverse time, inverse tokens) and three from the protocols (ownership clarity, handoff inspectability, useful parallelism). The two panels separate planned-handoff modes from independent-then-reconcile modes.

\textbf{The hardest projects favor modes that turn long requirement chains into explicit checks.} Online live teaching and AI image restoration are the two lowest-scoring full-grid projects (average best scores 67.1 and 69.1). They are hard not only because they need multimedia support, but because many user-visible functions must compose: live teaching has 24 functional items across six groups, and image restoration chains upload/storage, ordered enhancement calls, progress, before-after comparison, exhibit layout, search, and moderation. QA-first is strongest on image restoration (78.4, +9.3 over the project average) and above average on live teaching (69.4), because it turns requirement documents into acceptance targets before implementation spreads across API calls, realtime state, and UI flows matching \textsc{TAgent} validation.

\textbf{Ownership is most useful when multiple agents share one requirement.} Feature squad is a top-three mode on five projects, reaching 82.4 on enterprise asset management and 92.7 on e-commerce. An interaction chain such as approving a user or publishing an exhibit lives in neither frontend nor backend alone---it must update persistent state, call the right APIs, refresh the visible page, and enforce role permissions. If agents split only by technical layer, no one owns the whole scenario after \textsc{TAgent} reports an unmet requirement; feature squad instead gives one agent the complete vertical path. This is why it stays competitive across long-path projects, while swarming and open-source review often spend later rounds reconciling overlapping changes.

\textbf{Large repair gains can hide weak first rounds.} The R1-score axis separates modes: QA-first and pipeline start higher because their plans expose tests or staged deliverables early, while open-source review starts low and relies on later correction. Feature squad stays most balanced (high score, gain, consistency, and speed). Competitive and open-source review post large R1-to-R3 gains, but their weaker score or cost shows a big gain can mean a weak first round rather than a better policy.

\begin{figure*}[!t]
\centering
\includegraphics[width=\textwidth]{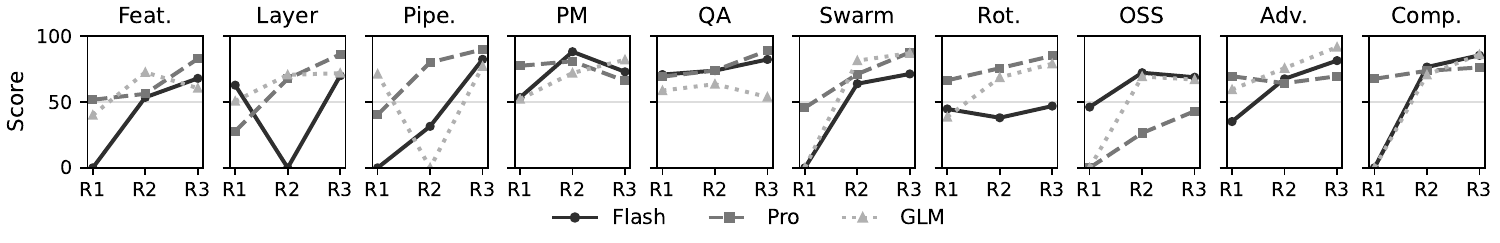}
\caption{Per-round trajectories of instant messaging scores over R1--R3.}
\label{fig:round_trajectories}
\end{figure*}

\subsection{\textsc{TAgent} Feedback Drives Refinement}

\textbf{\textsc{TAgent} feedback turns weak first attempts into measurable refinement.} Each round starts from the previous round's deployed codebase, and \textsc{TAgent} reports requirement-level failures to repair. Across the 600 adjacent transitions in the 300 full-grid runs, 82.0\% show score improvement, and 94.7\% of runs finish R3 above R1. These numbers make R1 a one-shot baseline and R3 the feedback-conditioned result: the benchmark measures whether agents can use concrete evaluator evidence, not merely whether they can retry.

\textbf{Three rounds capture most of the remaining gain.} 73.3\% of R2-to-R3 steps gain under 20 points (median +7.8), so we stop at three rounds to capture the main refinement effect without an open-ended budget search. GLM-5.2 shows the largest late gain when not cut off early, but the aggregate curve suggests further rounds would mostly study tail recoveries rather than the typical loop.

\textbf{Regression is a distinct failure mode.} Two mechanisms explain the low stability axis in Figure~\ref{fig:mode_radar} and the round-level drops in Table~\ref{tab:eval_models}. First, the per-round time budget can truncate implementation before the repository is coherent; this is most visible for GLM-5.2, whose many first rounds hit the 90-minute cap with unfinished handlers or missing routes. Second, a later repair can fix one requirement while breaking another: in the DeepSeek v4 Pro PM-oversight run on instant messaging, round 3 fixes profile editing and API docs but regresses user search, friend application, message sending, unread counts, and several group flows, dropping the score from 80.9 to 66.0. \sys thus reproduces the software-engineering failure modes a round-by-round benchmark should expose.

\subsection{A General Failure Taxonomy}

Regressions explain lost \emph{stability}, but not why even the best round rarely reaches 100. We therefore label every unmet \textsc{TAgent} rubric check by root cause, yielding 1{,}440 agent-attributable failing checks (a further 80, 5.3\%, are grader-side infrastructure errors we exclude); Table~\ref{tab:failure_taxonomy} gives each mode's share of failing checks and the fraction of runs it touches. 86\% of failed items still retain partial credit---only 14\% score zero---so teams almost always ship a running-but-incomplete system rather than a crashed one, and no single cause dominates.

\begin{table}[!t]
\centering
\small
\setlength{\tabcolsep}{1.4mm}
\renewcommand{\arraystretch}{0.9}
\begin{tabular}{@{}lcc@{}}
\toprule
\textbf{Failure mode} & \textbf{\% checks} & \textbf{\% runs} \\
\midrule
Incomplete logic \& edge cases          & 32.2 & 88 \\
Missing / unimplemented feature         & 23.8 & 75 \\
Auth, session \& permission             & 11.9 & 46 \\
Prerequisite cascade                    & 11.0 & 39 \\
Security \& transport (deploy ceiling)  & \phantom{0}8.5 & 78 \\
Other functional defect                 & \phantom{0}6.1 & 38 \\
Documentation gaps                      & \phantom{0}3.9 & 26 \\
Integration \& UI wiring                & \phantom{0}2.6 & 22 \\
\bottomrule
\end{tabular}
\caption{General failure taxonomy over 1{,}440 failing \textsc{TAgent} rubric checks. ``\% checks'' is the share of all failing checks; ``\% runs'' is the fraction of runs with at least one such failure.}
\label{tab:failure_taxonomy}
\end{table}

\textbf{Functional shortfall dominates, not crashes.} Missing features (23.8\%) and incomplete logic or edge cases (32.2\%) together account for 56\% of failing checks and a similar share of lost weight. The typical defect is an endpoint that exists but skips a validation, a cleanup, or a secondary path---account deletion that invalidates the token but never cascades to messages, or a history filter that returns empty on a valid range---exactly the requirements a document-driven grader surfaces but a self-reported ``done'' would hide.

\textbf{One broken prerequisite poisons many downstream checks.} Authentication and permission failures (11.9\%) and prerequisite cascades (11.0\%) are together 23\% of all failures. When login, token refresh, or a gating step such as friend-request breaks, every dependent scenario becomes untestable: a single broken prerequisite auto-skipped up to 14 downstream checks in one run, and cascades appear in 39\% of runs. This is the multi-agent-specific failure mode---a contract broken by one owner silently zeroes another owner's correct work---and why requirement-level feedback with explicit dependencies beats a scalar score.

\textbf{Part of the ceiling is a fixed deployment artifact.} Security and transport items (8.5\% of checks) recur in nearly 80\% of runs because they depend on successful integration with \textsc{XDeploy}, the deployment layer that supplies TLS, HTTPS redirects, and secure WebSockets; deployment is carried out through \textsc{XDeploy} rather than locally, so these are a near-constant deduction from unsuccessful integration rather than a coding error. With documentation (3.9\%) and UI-wiring defects (2.6\%) minor, the benchmark is bottlenecked not on raw code generation but on integration completeness and cross-agent contracts.

\subsection{Key Findings and Insights}

\textbf{Topology is the strongest effect, and it is coupled with the model.} DeepSeek v4 Pro swings from 89.9 under pipeline to 43.0 under open-source review on instant messaging, and across the 10-mode average QA-first and rotation tie highest at 83.3 while open-source review drops to 73.5. The same policy also behaves differently across models: Flash improves fast but plateaus on structurally incomplete first drafts, Pro is the stable middle ground, and GLM-5.2 recovers sharply once implementation time catches up with its slow decoding. Early inspection helps most---QA-first converts requirements into acceptance targets before implementation spreads across APIs, state, and UI, whereas late review must reconcile already-divergent changes.

\textbf{The score frontier is not the cost frontier.} GLM-5.2 reaches the highest full-grid average best score but at far higher time and cost than the DeepSeek models, and Flash is cheapest in every Flash-vs-Pro pair despite using more raw tokens in 34 of 100 configurations. Because several models reach a competitive score with much less spend, the practical choice is the best score-per-budget point, and the benchmark separates token volume from actual budget pressure.

\textbf{Coordination policy is the control surface.} The largest score swings come from how work is divided and merged, not from raw model identity: parallelism helps only when feedback still maps to an accountable owner, or extra agents turn a requirement failure into a negotiation over whose change survives. The metric signatures make this legible---a low-score, low-time run signals early deployment failure, while a high-token, medium-score run signals repeated re-planning or repair of the wrong layer. Future agentic systems should therefore learn when to parallelize, when to serialize, and when to stop a branch before it costs more than its expected gain---a sensitivity \sys makes measurable.

\section*{Limitations}

\noindent\textbf{Randomness and scale.} To mitigate LLM stochasticity we run three parallel trials for the instant-messaging token/time evaluation and rerun other projects only when experts flag a test as abnormal (\textit{e.g.,} API failures). The evaluation scale is nonetheless bounded by the high test budget.

\noindent\textbf{\textsc{TAgent} and the task mix are scoped.} \textsc{TAgent} can miss some edge defects, and \sys centers web apps. On student projects the \textsc{TAgent} score attains a Spearman correlation of 0.87 with TA scores, indicating high agreement. We keep the scope to real-world apps and leave directions such as autoresearch to others~\cite{xiong2026autoresearchbenchbenchmarkingaiagents,du2025deepresearchbenchcomprehensivebenchmark}.

\noindent\textbf{\sys does not cover the full software lifecycle.} Incident response and long-lived maintenance also matter, but \sys focuses on collaboration topology, iteration, and cost during the coding stage; we leave the rest to future work.

\section{Conclusion}
\sys proposes a 100 project--collaboration mode benchmark that examines project, model, topology, time, token usage cost, and feedback rounds together. \sys features an automatic feedback and revision pipeline for requirement documents without API specification details. Evaluations show that modern large language models can build substantial software but remain strongly shaped by orchestration, and reveal that future agents should treat collaboration mode, model choice, and stopping policy as runtime decisions rather than fixed defaults.

\bibliography{aaai2027}

@inproceedings{multiagentbench,
  title={Multiagentbench: Evaluating the collaboration and competition of llm agents},
  author={Zhu, Kunlun and Du, Hongyi and Hong, Zhaochen and Yang, Xiaocheng and Guo, Shuyi and Wang, Daisy Zhe and Wang, Zhenhailong and Qian, Cheng and Tang, Robert and Ji, Heng and others},
  booktitle={Proceedings of the 63rd Annual Meeting of the Association for Computational Linguistics (Volume 1: Long Papers)},
  pages={8580--8622},
  year={2025}
}

@inproceedings{swebench,
  title={Swe-bench: Can language models resolve real-world github issues?},
  author={Jimenez, Carlos E and Yang, John and Wettig, Alexander and Yao, Shunyu and Pei, Kexin and Press, Ofir and Narasimhan, Karthik},
  booktitle={International Conference on Learning Representations},
  volume={2024},
  pages={54107--54157},
  year={2024}
}

@inproceedings{sweagent,
  title={Swe-agent: Agent-computer interfaces enable automated software engineering},
  author={Yang, John and Jimenez, Carlos E and Wettig, Alexander and Lieret, Kilian and Yao, Shunyu and Narasimhan, Karthik R and Press, Ofir},
  booktitle={The Thirty-eighth Annual Conference on Neural Information Processing Systems},
  year={2024}
}

@article{agentless,
author = {Xia, Chunqiu Steven and Deng, Yinlin and Dunn, Soren and Zhang, Lingming},
title = {Demystifying LLM-Based Software Engineering Agents},
year = {2025},
issue_date = {July 2025},
publisher = {Association for Computing Machinery},
address = {New York, NY, USA},
volume = {2},
number = {FSE},
url = {https://doi.org/10.1145/3715754},
doi = {10.1145/3715754},
journal = {Proc. ACM Softw. Eng.},
month = jun,
articleno = {FSE037},
numpages = {24}
}

@inproceedings{autocoderover,
  title={Autocoderover: Autonomous program improvement},
  author={Zhang, Yuntong and Ruan, Haifeng and Fan, Zhiyu and Roychoudhury, Abhik},
  booktitle={Proceedings of the 33rd ACM SIGSOFT International Symposium on Software Testing and Analysis},
  pages={1592--1604},
  year={2024}
}

@inproceedings{openhands,
  title={Openhands: An open platform for ai software developers as generalist agents},
  author={Wang, Xingyao and Li, Boxuan and Song, Yufan and Xu, Frank F and Tang, Xiangru and Zhuge, Mingchen and Pan, Jiayi and Song, Yueqi and Li, Bowen and Singh, Jaskirat and others},
  booktitle={International Conference on Learning Representations},
  volume={2025},
  pages={65882--65919},
  year={2025}
}

@inproceedings{repobench,
 author = {Liu, Tianyang and Xu, Canwen and McAuley, Julian },
 booktitle = {International Conference on Learning Representations},
 editor = {B. Kim and Y. Yue and S. Chaudhuri and K. Fragkiadaki and M. Khan and Y. Sun},
 pages = {47832--47850},
 title = {RepoBench: Benchmarking Repository-Level Code Auto-Completion Systems},
 url = {https://proceedings.iclr.cc/paper_files/paper/2024/file/d191ba4c8923ed8fd8935b7c98658b5f-Paper-Conference.pdf},
 volume = {2024},
 year = {2024}
}

@misc{humaneval,
      title={Evaluating Large Language Models Trained on Code}, 
      author={Mark Chen and Jerry Tworek and Heewoo Jun and Qiming Yuan and Henrique Ponde de Oliveira Pinto and Jared Kaplan and Harri Edwards and Yuri Burda and Nicholas Joseph and Greg Brockman and Alex Ray and Raul Puri and Gretchen Krueger and Michael Petrov and Heidy Khlaaf and Girish Sastry and Pamela Mishkin and Brooke Chan and Scott Gray and Nick Ryder and Mikhail Pavlov and Alethea Power and Lukasz Kaiser and Mohammad Bavarian and Clemens Winter and Philippe Tillet and Felipe Petroski Such and Dave Cummings and Matthias Plappert and Fotios Chantzis and Elizabeth Barnes and Ariel Herbert-Voss and William Hebgen Guss and Alex Nichol and Alex Paino and Nikolas Tezak and Jie Tang and Igor Babuschkin and Suchir Balaji and Shantanu Jain and William Saunders and Christopher Hesse and Andrew N. Carr and Jan Leike and Josh Achiam and Vedant Misra and Evan Morikawa and Alec Radford and Matthew Knight and Miles Brundage and Mira Murati and Katie Mayer and Peter Welinder and Bob McGrew and Dario Amodei and Sam McCandlish and Ilya Sutskever and Wojciech Zaremba},
      year={2021},
      eprint={2107.03374},
      archivePrefix={arXiv},
      primaryClass={cs.LG},
      url={https://arxiv.org/abs/2107.03374}, 
}

@misc{mbpp,
      title={Program Synthesis with Large Language Models}, 
      author={Jacob Austin and Augustus Odena and Maxwell Nye and Maarten Bosma and Henryk Michalewski and David Dohan and Ellen Jiang and Carrie Cai and Michael Terry and Quoc Le and Charles Sutton},
      year={2021},
      eprint={2108.07732},
      archivePrefix={arXiv},
      primaryClass={cs.PL},
      url={https://arxiv.org/abs/2108.07732}, 
}

@article{theagentcompany,
  title={Theagentcompany: benchmarking llm agents on consequential real world tasks},
  author={Xu, Frank Fangzheng and Song, Yufan and Li, Boxuan and Tang, Yuxuan and Jain, Kritanjali and Bao, Mengxue and Wang, Zora and Zhou, Xuhui and Guo, Zhitong and Cao, Murong and others},
  journal={Advances in Neural Information Processing Systems},
  volume={38},
  year={2026}
}

@misc{agentsnet,
      title={AgentsNet: Coordination and Collaborative Reasoning in Multi-Agent LLMs}, 
      author={Florian Grötschla and Luis Müller and Jan Tönshoff and Mikhail Galkin and Bryan Perozzi},
      year={2025},
      eprint={2507.08616},
      archivePrefix={arXiv},
      primaryClass={cs.MA},
      url={https://arxiv.org/abs/2507.08616}, 
}

@inproceedings{clouddevbench,
author = {Guan, Tian},
title = {A Multi-Agent Coding Assistant for Cloud-Native Development: From Requirements to Deployable Microservices},
year = {2026},
isbn = {9798400719981},
publisher = {Association for Computing Machinery},
address = {New York, NY, USA},
url = {https://doi.org/10.1145/3795154.3795362},
doi = {10.1145/3795154.3795362},
booktitle = {Proceedings of the 2025 6th International Conference on Computer Science and Management Technology},
pages = {1285–1290},
numpages = {6},
location = {
},
series = {ICCSMT '25}
}

@misc{deepswe,
      title={DeepSWE: Measuring Frontier Coding Agents on Original, Long-Horizon Engineering Tasks}, 
      author={Wenqi Huang and Charley Lee and Leonard Tng and Serena Ge},
      year={2026},
      eprint={2607.07946},
      archivePrefix={arXiv},
      primaryClass={cs.SE},
      url={https://arxiv.org/abs/2607.07946}, 
}

@misc{frontiercode,
  title        = "Introducing FrontierCode",
  author       = {Lu, Eric and Pan, Ben and Birlikci, Deniz and Lee, Sam and Wang, Ray and Choudhury, Rohan and Ma, Fermi and Qin, TC and Baronio, Carlo and Alberti, Silas and others},
  howpublished = "\url{https://cognition.com/blog/frontier-code}",
  year         = 2026,
  note         = "Accessed: 2026-07-13"
}

@misc{anthropic_claude48,
  title        = "Introducing Claude Opus 4.8",
  author       = "{Anthropic}",
  howpublished = "\url{https://www.anthropic.com/news/claude-opus-4-8}",
  year         = 2026,
  note         = "Accessed: 2026-07-20"
}

@misc{openai_gpt55,
  title        = "Introducing GPT-5.5",
  author       = "{OpenAI}",
  howpublished = "\url{https://openai.com/index/introducing-gpt-5-5/}",
  year         = 2026,
  note         = "Accessed: 2026-07-20"
}

@misc{deepseek_v4,
      title={DeepSeek-V4: Towards Highly Efficient Million-Token Context Intelligence}, 
      author={DeepSeek-AI and Anyi Xu and Bangcai Lin and Bing Xue and Bingxuan Wang and Bingzheng Xu and Bochao Wu and Bowei Zhang and Chaofan Lin and Chen Dong and Chenchen Ling and Chengda Lu and Chenggang Zhao and Chengqi Deng and Chengyu Hou and Chenhao Xu and Chenze Shao and Chong Ruan and Conner Sun and Damai Dai and Daya Guo and Dejian Yang and Deli Chen and Donghao Li and Dongjie Ji and Erhang Li and Fang Wei and Fangyun Lin and Fangzhou Yuan and Feiyu Xia and Fucong Dai and Guangbo Hao and Guanting Chen and Guoai Cao and Guolai Meng and Guowei Li and Han Yu and Han Zhang and Hanwei Xu and Hao Li and Haofen Liang and Haoling Zhang and Haoming Luo and Haoran Wei and Haotian Yuan and Haowei Zhang and Haowen Luo and Haoyu Chen and Haozhe Ji and Hengqing Zhang and Honghui Ding and Hongxuan Tang and Huanqi Cao and Huazuo Gao and Hui Qu and Hui Zeng and J Yang and JQ Zhu and Jia Luo and Jia Song and Jia Yu and Jialiang Huang and Jialu Cai and Jian Liang and Jiangting Zhou and Jiasheng Ye and Jiashi Li and Jiaxin Xu and Jiewen Hu and Jieyu Yang and Jin Chen and Jin Yan and Jingchang Chen and Jingli Zhou and Jingting Xiang and Jingyang Yuan and Jingyuan Cheng and Jingzi Zhou and Jinhua Zhu and Jiping Yu and Joseph Sun and Jun Ran and Junguang Jiang and Junjie Qiu and Junlong Li and Junmin Zheng and Junxiao Song and Kai Dong and Kaige Gao and Kang Guan and Kexing Zhou and Kezhao Huang and Kuai Yu and Lean Wang and Lecong Zhang and Lei Wang and Leyi Xia and Li Zhang and Liang Zhao and Lihua Guo and Lingxiao Luo and Linwang Ma and Linyan Zhu and Litong Wang and Liyu Cai and Liyue Zhang and Longhao Chen and MS Di and MY Xu and Max Mei and Miaojun Wang and Mingchuan Zhang and Minghua Zhang and Minghui Tang and Mingming Li and Mingxu Zhou and Minmin Han and Ning Wang and Panpan Huang and Panpan Wang and Peixin Cong and Peiyi Wang and Peng Zhang and Qiancheng Wang and Qihao Zhu and Qingyang Li and Qinyu Chen and Qiushi Du and Qiwei Jiang and Rui Tian and Ruifan Xu and Ruijie Lu and Ruiling Xu and Ruiqi Ge and Ruisong Zhang and Ruizhe Pan and Runji Wang and Runqian Chen and Runqiu Yin and Runxin Xu and Ruomeng Shen and Ruoyu Zhang and Ruyi Chen and SH Liu and Shanghao Lu and Shangmian Sun and Shangyan Zhou and Shanhuang Chen and Shaofei Cai and Shaoheng Nie and Shaoqing Wu and Shaoyuan Chen and Shengding Hu and Shengyu Liu and Shiqiang Hu and Shirong Ma and Shiyu Wang and Shuiping Yu and Shunfeng Zhou and Shuting Pan and Shuying Yu and Songyang Zhou and Tao Ni and Tao Yun and Tian Jin and Tian Pei and Tian Ye and Tianle Lin and Tianran Ji and Tianyi Cui and Tianyuan Yue and Tingting Yu and Tun Wang and W Zhang and WL Xiao and Wangding Zeng and Wei An and Weilin Zhao and Wen Liu and Wenfeng Liang and Wenjie Pang and Wenjing Luo and Wenjing Yao and Wenjun Gao and Wenkai Yang and Wenlve Huang and Wenqing Hou and Wentao Zhang and Wenting Ma and Xi Gao and Xiang He and Xiangwen Wang and Xianzu Wang and Xiao Bi and Xiaodong Liu and Xiaohan Wang and Xiaokang Chen and Xiaokang Zhang and Xiaotao Nie and Xiaowen Sun and Xiaoxiang Wang and Xin Cheng and Xin Liu and Xin Xie and Xingchao Liu and Xingchen Liu and Xingkai Yu and Xingyou Li and Xinyu Yang and Xinyu Zhang and Xu Chen and Xuanyu Wang and Xuecheng Su and Xueyin Chen and Xuheng Lin and Xuwei Fu and YC Yan and YQ Wang and YW Ma and Yanfeng Luo and Yang Zhang and Yanhong Xu and Yanru Ma and Yanwen Huang and Yao Li and Yao Li and Yao Xu and Yao Zhao and Yaofeng Sun and Yaohui Wang and Yi Qian and Yi Shao and Yi Yu and Yichao Zhang and Yifan Ding and Yifan Shi and Yijia Wu and Yiliang Xiong and Yiling Ma and Ying He and Ying Tang and Ying Zhou and Yingjia Luo and Yinmin Zhong and Yishi Piao and Yisong Wang and Yixiang Zhang and Yixiao Chen and Yixuan Tan and Yixuan Wei and Yiyang Ma and Yiyuan Liu and Yonglun Yang and Yongqiang Guo and Yongtong Wu and Yu Wu and YuKun Li and Yuan Cheng and Yuan Ou and Yuanfan Xu and Yuanhao Li and Yuduan Wang and Yuehan Yang and Yuer Xu and Yuhan Wu and Yuhao Meng and Yuheng Zou and Yukun Zha and Yunfan Xiong and Yupeng Chen and Yuping Lin and Yuqian Cao and Yuqian Wang and Yushun Zhang and Yuting Yan and Yutong Lin and Yuxian Gu and Yuxiang Luo and Yuxiang You and Yuxuan Liu and Yuxuan Zhou and Yuyang Zhou and Yuzhen Huang and ZF Wu and Zehao Wang and Zehua Zhao and Zehui Ren and Zekai Zhang and Zhangli Sha and Zhe Fu and Zhe Ju and Zhean Xu and Zhenda Xie and Zhengyan Zhang and Zheren Gao and Zhewen Hao and Zhibin Gou and Zhicheng Ma and Zhigang Yan and Zhihong Shao and Zhixian Huang and Zhixuan Chen and Zhiyu Wu and Zhizhou Ren and Zhongyu Wu and Zhuoshu Li and Zhuping Zhang and Zian Xu and Zihao Wang and Zihua Qu and Zihui Gu and Zijia Zhu and Zilin Li and Zipeng Zhang and Ziwei Xie and Ziyi Gao and Ziyi Wan and Zizheng Pan and Zongqing Yao},
      year={2026},
      eprint={2606.19348},
      archivePrefix={arXiv},
      primaryClass={cs.CL},
      url={https://arxiv.org/abs/2606.19348}, 
}

@misc{glm52,
  title        = "GLM-5.2: Built for Long-Horizon Tasks",
  author       = "{Z.AI}",
  howpublished = "\url{https://z.ai/blog/glm-5.2}",
  year         = 2026,
  note         = "Accessed: 2026-07-15"
}

@misc{qwen36flash,
    title = {{Qwen3.6-35B-A3B}: Agentic Coding Power, Now Open to All},
    author = {{Qwen Team}},
    year = {2026},
    month = {April},
    url = {https://qwen.ai/blog?id=qwen3.6-35b-a3b}
}

@misc{anthropic_opus48_pricing,
  title        = "Claude API Pricing",
  author       = "{Anthropic}",
  howpublished = "\url{https://platform.claude.com/docs/en/about-claude/pricing}",
  year         = 2026,
  note         = "Accessed: 2026-07-20"
}

@misc{openai_gpt55_pricing,
  title        = "OpenAI API Pricing",
  author       = "{OpenAI}",
  howpublished = "\url{https://developers.openai.com/api/docs/pricing}",
  year         = 2026,
  note         = "Accessed: 2026-07-20"
}

@misc{deepseek_v4_pricing,
  title        = "Models and Pricing",
  author       = "{DeepSeek}",
  howpublished = "\url{https://api-docs.deepseek.com/quick_start/pricing/}",
  year         = 2026,
  note         = "Accessed: 2026-07-18"
}

@misc{zai_pricing,
  title        = "{Z.AI} Pricing",
  author       = "{Z.AI}",
  howpublished = "\url{https://docs.z.ai/guides/overview/pricing}",
  year         = 2026,
  note         = "Accessed: 2026-07-15"
}

@misc{qwen_pricing,
  title        = "Alibaba Cloud Model Studio: Model Inference Pricing",
  author       = "{Alibaba Cloud}",
  howpublished = "\url{https://www.alibabacloud.com/help/en/model-studio/model-pricing}",
  year         = 2026,
  note         = "Accessed: 2026-07-17"
}

@book{sonarqube,
  title={SonarQube in action},
  author={Campbell, G Ann and Papapetrou, Patroklos P},
  year={2013},
  publisher={Manning Publications Co.}
}

@misc{xiong2026autoresearchbenchbenchmarkingaiagents,
      title={AutoResearchBench: Benchmarking AI Agents on Complex Scientific Literature Discovery}, 
      author={Lei Xiong and Kun Luo and Ziyi Xia and Wenbo Zhang and Jin-Ge Yao and Zheng Liu and Jingying Shao and Jianlyu Chen and Hongjin Qian and Xi Yang and Qian Yu and Hao Li and Chen Yue and Xiaan Du and Yuyang Wang and Yesheng Liu and Haiyu Xu and Zhicheng Dou},
      year={2026},
      eprint={2604.25256},
      archivePrefix={arXiv},
      primaryClass={cs.AI},
      url={https://arxiv.org/abs/2604.25256}, 
}

@misc{du2025deepresearchbenchcomprehensivebenchmark,
      title={DeepResearch Bench: A Comprehensive Benchmark for Deep Research Agents}, 
      author={Mingxuan Du and Benfeng Xu and Chiwei Zhu and Xiaorui Wang and Zhendong Mao},
      year={2025},
      eprint={2506.11763},
      archivePrefix={arXiv},
      primaryClass={cs.CL},
      url={https://arxiv.org/abs/2506.11763}, 
}


\newpage
\section*{Appendix}
\section{Benchmark Assets}

We will release the benchmark in \url{https://github.com/robinren03/MSEval}.

\section{The Ten Projects}

\sys draws its ten projects from real software-engineering course
capstones spanning messaging, CRUD-and-workflow business systems,
transactional commerce, retrieval, analytics, access control, and
realtime multimedia. Each ships a hierarchical requirement document that
compiles to a weighted 100-point rubric (typically 35--40 system, 50--55
functional, 10 documentation). Table~\ref{tab:projects} lists them; only
the three full-grid models (DeepSeek v4 Flash/Pro, GLM-5.2) run all ten,
while Claude Opus 4.8, GPT-5.5, and Qwen3.6-Flash run only the
instant-messaging reference project P00 (main-paper results table).

\begin{table}[t]
\centering
\small
\setlength{\tabcolsep}{1.4mm}
\renewcommand{\arraystretch}{1.05}
\begin{tabular}{@{}llc@{}}
\toprule
\textbf{ID} & \textbf{Project (domain)} & \textbf{Scope} \\
\midrule
P00 & Instant messaging          & 56 checks \\
P01 & Enterprise asset management & long-path \\
P02 & Crowdsourcing platform      & multi-role \\
P03 & Requirement tracking        & workflow \\
P04 & AI image restoration        & multimedia \\
P05 & E-commerce trading          & transactional \\
P06 & Influencer analytics        & data/vis \\
P07 & Permission platform         & auth-heavy \\
P08 & News search                 & retrieval \\
P09 & Online live teaching        & realtime A/V \\
\bottomrule
\end{tabular}
\caption{The ten projects. ``Scope'' names the dominant difficulty; e.g.\
online live teaching has 24 functional items across six groups and AI
image restoration chains upload/storage, ordered enhancement calls,
progress, before/after comparison, exhibit layout, search, and
moderation. Every project ships a hierarchical requirement document that
compiles to a weighted rubric (P00 yields 56 checks: 7 system, 46
functional, 3 documentation).}
\label{tab:projects}
\end{table}

\noindent Each project below is annotated with its full-grid difficulty -- the mean best score over the three full-grid models (DeepSeek v4 Flash/Pro and GLM-5.2) across the ten collaboration modes -- and the collaboration mode that tops it.

\begin{itemize}
  \item \textbf{P00 Instant messaging} (mean 77.2; best PM oversight 83.7). An iMessage-style app for private and group chat with a session-list/conversation-pane UI, serving ordinary users and group owners/admins. Its 56-check rubric splits 40 system (usability, security, sub-200~ms realtime delivery, data integrity) / 50 functional / 10 docs, with friend- and group-prerequisite chains; it is the reference project run by all five models.
  \item \textbf{P01 Enterprise asset management} (mean 70.3; best pipeline 83.0; GLM-5.2 leads at 76.3). Full asset lifecycle -- intake, lending, transfer, retirement, maintenance -- with multi-tenant business-entity isolation, hierarchical category trees, and approval workflows across four role tiers. A single operation must mutate persistent state, enforce tenant isolation, and append an audit log, so the layer split collapses here (57.2, its worst project).
  \item \textbf{P02 Crowdsourcing platform} (mean 74.7; best PM oversight 80.9). A data-labeling marketplace spanning publisher, worker, and reviewer roles over a long chain: create/configure task $\to$ upload data $\to$ dispatch $\to$ claim $\to$ annotate $\to$ review $\to$ settle $\to$ export. Anti-cheating and credit scoring ride on top of every stage, keeping the whole pipeline coupled.
  \item \textbf{P03 Requirement tracking} (mean 78.1; best QA-first 91.7; DeepSeek v4 Pro leads at 86.4). A Jira-like tool that decomposes initial requirements into functional requirements, plans iterations, and analyzes Git-commit traceability across several services. QA-first dominates because its acceptance targets map directly onto the traceability rubric.
  \item \textbf{P04 AI image restoration} (mean 69.1; best QA-first 78.4; GLM-5.2 leads at 85.6). The second-hardest project: museum photo restoration via online AI APIs plus curated exhibition rooms with before/after comparison, browse/search, and likes/comments, serving curators and visitors. The multimedia chain -- upload/storage $\to$ ordered enhancement calls $\to$ progress $\to$ comparison $\to$ exhibit -- punishes weak models (Flash 56.9) and rewards GLM-5.2's long-horizon planning (85.6).
  \item \textbf{P05 E-commerce trading} (mean 89.9; best pipeline 96.2). A self-operated marketplace with buyer-side browse, balance top-up, and order flows plus merchant-side catalog, discount, and order management. Transactional consistency with clean stage boundaries makes the serial pipeline strongest.
  \item \textbf{P06 Influencer analytics} (mean 90.5; best competitive 96.5). A creator dashboard over a bundled dataset of per-day view/like/comment deltas, offering overview, per-work analysis, fan-activity trends, and background refresh. Being largely read-only over a fixed dataset places it in the top scoring tier.
  \item \textbf{P07 Permission platform} (mean 91.0; best layer specialists 96.3; GLM-5.2 leads at 94.8). An RBAC platform covering apps, permission types, resources, roles, user authorization, and an external permission-query API with tenant isolation. Its clean split between a configuration UI and an auth API is the one case where layer specialists win.
  \item \textbf{P08 News search} (mean 92.8; best QA-first 99.1; DeepSeek v4 Pro leads at 97.7). The easiest project: a full-text search engine -- crawl, index, retrieve, rank, present -- plus user interest tags. The well-bounded retrieval pipeline yields the highest scores in the grid.
  \item \textbf{P09 Online live teaching} (mean 67.1; best feature squad 75.1). The hardest project: a WebRTC live classroom for teacher, TA, and student roles with streaming, teaching resources, room management, in-class chat, and quizzes -- 24 functional items across six groups. Only vertical ownership (feature squad) keeps the composed realtime features coherent.
\end{itemize}

\begin{table}[t]
\centering
\small
\setlength{\tabcolsep}{1.4mm}
\renewcommand{\arraystretch}{1.0}
\begin{tabular}{@{}lccccl@{}}
\toprule
\textbf{ID} & \textbf{Flash} & \textbf{Pro} & \textbf{GLM} & \textbf{Avg} & \textbf{Best mode} \\
\midrule
P00 & 74.8 & 79.0 & 78.0 & \textbf{77.2} & PM (83.7) \\
P01 & 65.1 & 69.5 & 76.3 & \textbf{70.3} & Pipe (83.0) \\
P02 & 69.1 & 75.2 & 80.0 & \textbf{74.7} & PM (80.9) \\
P03 & 76.0 & 86.4 & 72.0 & \textbf{78.1} & QA (91.7) \\
P04 & 56.9 & 64.8 & 85.6 & \textbf{69.1} & QA (78.4) \\
P05 & 86.5 & 93.1 & 90.2 & \textbf{89.9} & Pipe (96.2) \\
P06 & 86.1 & 93.8 & 91.6 & \textbf{90.5} & Comp (96.5) \\
P07 & 84.5 & 93.6 & 94.8 & \textbf{91.0} & Layer (96.3) \\
P08 & 86.7 & 97.7 & 94.1 & \textbf{92.8} & QA (99.1) \\
P09 & 56.2 & 72.1 & 73.0 & \textbf{67.1} & Feat (75.1) \\
\bottomrule
\end{tabular}
\caption{Full-grid best-score results per project.
\textbf{Flash}/\textbf{Pro}/\textbf{GLM} are each model's mean best score
over the ten modes; \textbf{Avg} is the mean over the three models;
\textbf{Best mode} is the top mode on that project with its mean best
score. Claude Opus 4.8, GPT-5.5, and Qwen3.6-Flash run only P00 and
appear in the main-paper results table.}
\label{tab:proj_perf}
\end{table}

\noindent\textbf{Difficulty tracks requirement composition, not raw feature count.} The two hardest projects -- live teaching (67.1) and image restoration (69.1) -- together with enterprise asset management (70.3) force many user-visible functions to compose over shared realtime or transactional state, whereas the easiest -- news search (92.8), permission (91.0), and analytics (90.5) -- decompose into well-bounded, largely independent requirements. Across Table~\ref{tab:proj_perf}, DeepSeek v4 Pro is the steadiest full-grid model, posting the highest score on five of ten projects; GLM-5.2 wins exactly the hardest multimedia and access-control projects (P04, P07), consistent with its long-horizon strength; and DeepSeek v4 Flash keeps pace on the easy retrieval/CRUD projects but falls 15--30 points behind on multimedia and realtime (P04, P09).

\section{The Ten Collaboration Modes}

Every mode is an instance of the same template with four concrete
objects: an \emph{ownership map} (who owns which part of the repository),
an \emph{activation schedule} (who is awake in which round), a
\emph{decision rule} (how conflicts resolve), and a \emph{required handoff
artifact} (what must exist before a round can be accepted). Teams are
four agents by default (\texttt{dev-1}\,\dots\,\texttt{dev-4}, or role
names such as architect/backend/frontend/QA for staged modes).
Table~\ref{tab:modes} gives the ten definitions, grouped by how they
coordinate; the grouping matches the two radar panels in the main paper
(serialized handoff vs.\ parallel contention).

\begin{table*}[t]
\centering
\small
\setlength{\tabcolsep}{1.6mm}
\renewcommand{\arraystretch}{1.15}
\begin{tabular}{@{}p{2.05cm}p{3.0cm}p{3.0cm}p{3.0cm}p{3.6cm}@{}}
\toprule
\textbf{Mode} & \textbf{Ownership map} & \textbf{Activation schedule} & \textbf{Decision rule} & \textbf{Required handoff artifact} \\
\midrule
\multicolumn{5}{@{}l}{\textit{Serialized handoff (planned coordination)}}\\
\midrule
Feature squad & One agent owns a full vertical user-facing module (UI+API+state). & All agents active in parallel every round. & Module owner is authoritative for their scenario end to end. & A working vertical slice per module, integrated on the shared port. \\
Layer specialists & Split by technical layer (frontend vs.\ backend). & Parallel, synchronizing at layer boundaries. & API contract is the arbiter between layers. & An agreed API contract plus both layers built against it. \\
Pipeline & Disjoint stages: architect $\to$ backend $\to$ frontend $\to$ QA. & Almost serial; each stage activates the next. & Upstream stage output is frozen for downstream. & A clean artifact boundary passed stage to stage (control condition). \\
QA-first & QA owns acceptance targets; devs own implementation. & Tests authored first, then implementation. & Acceptance tests are the source of truth. & An executable acceptance suite before feature code. \\
Rotation & Ownership rotates: agents switch roles each round. & All active; roles permuted per round. & Current round's role assignment governs. & A handoff note so the next role can continue the module. \\
\midrule
\multicolumn{5}{@{}l}{\textit{Parallel contention (independent-then-reconcile)}}\\
\midrule
PM oversight & A manager plans and reviews; devs implement. & Manager gates entry/exit of each round. & Manager adjudicates and assigns repairs. & A reviewed plan and a sign-off on each round's diff. \\
Swarming & No fixed ownership; tasks claimed dynamically. & All active; work pulled from a shared list. & First claimant owns the task until release. & A claimed-task ledger and merged result. \\
Open-source review & Contributors propose; reviewers must approve. & Parallel propose, then review rounds. & Change lands only on reviewer approval. & Reviewed change sets with approval trail. \\
Adversarial & Builders vs.\ an adversarial tester probing for defects. & Parallel build and attack. & Disputes resolved by reproducing the defect. & A defect report plus the fix that closes it. \\
Competitive & Redundant teams implement the same spec. & Parallel, independent implementations. & Best implementation is selected by score. & Two+ deployable candidates and a selection record. \\
\bottomrule
\end{tabular}
\caption{The ten collaboration modes as instances of the mode template.
Serialized-handoff modes pass work through explicit artifact boundaries;
parallel-contention modes maximize concurrency and must reconcile
overlapping changes. Pipeline is intentionally almost serial and serves
as a coordination-control condition.}
\label{tab:modes}
\end{table*}

\noindent Beyond the template objects in Table~\ref{tab:modes}, the ten modes differ in how they split the repository, who holds decision authority, and what most distinguishes each from its nearest neighbor.

\begin{itemize}
\item \textbf{Feature squad} splits the repository \emph{vertically} by user-facing module; four full-stack agents each own one module end to end (UI, API, state, tests) and align only through interface contracts, so decision authority is local to each owner. Unlike swarming it fixes module boundaries rather than leaving them fluid. Its signature is the most balanced across projects, and it is the only mode that tops the hardest project (P09, 75.1).
\item \textbf{Layer specialists} splits \emph{horizontally}, with two frontend and two backend agents who never cross layers and reconcile through a single API contract that holds the shared authority. It is the only technical-layer split. It wins when that contract is clean (P07, 96.3) but collapses when one requirement spans layers (P01, 57.2, the lowest full-grid cell).
\item \textbf{Pipeline} splits work \emph{temporally}: architect $\to$ backend $\to$ frontend $\to$ QA, each stage frozen before the next begins and holding authority only during its window. It is the only non-parallel mode and serves as our coordination-control condition. It is strongest on transactional/CRUD projects with clean stage boundaries (P05, 96.2; P01, 83.0).
\item \textbf{QA-first} is \emph{quality-driven}: QA authors an executable acceptance suite first and holds final say while developers chase the tests. Unlike adversarial testing the QA role is collaborative rather than attacking. It ties for the best cross-project mean and wins the requirement-heavy projects (P08, 99.1; P03, 91.7; P04, 78.4).
\item \textbf{Rotation} reassigns roles on a \emph{timer}: every round agents swap backend/frontend/QA/architect and continue the predecessor's work via a handoff note, so authority travels with the role rather than the agent. It is the only forced-rotation mode and the heaviest in dialogue turns. It ties QA-first for the best cross-project mean (83.3).
\item \textbf{PM oversight} is \emph{hierarchical}: a non-coding manager plans, assigns, and gates each round's entry and exit while developers only execute, concentrating authority at the top. Against open-source review's peer approval it is top-down command rather than horizontal consent. It leads on messaging and crowdsourcing (P00, 83.7; P02, 80.9).
\item \textbf{Swarming} applies \emph{no split}: four identical agents pull tasks from a shared list, the first claimant owning a task until release. It differs from feature squad by having no fixed boundaries and from rotation by not rotating roles. It sits mid-pack (79.3) and spends late rounds reconciling overlap.
\item \textbf{Open-source review} splits by \emph{permission}: a non-coding maintainer, two contributors, and a reviewer, with change landing only on approval, so authority rests in the approval gate. Against PM oversight it is horizontal peer review rather than top-down command. Its approval and reconciliation overhead make it consistently the weakest mode (73.5).
\item \textbf{Adversarial} runs \emph{red/blue}: a breaker probes for defects and can veto completion, with disputes settled by reproducing the defect. It is the only mode with a hostile role. It lands mid-pack (77.1) and is strongest where correctness edge cases dominate.
\item \textbf{Competitive} uses \emph{redundant contention}: two independent teams build the whole spec and the better is selected by score, placing authority in the scorer. It is the only two-team mode. It wins where a fixed dataset makes selection cheap (P06, 96.5) but is weakest on the hardest realtime project (P09, 59.1).
\end{itemize}

\begin{table*}[t]
\centering
\small
\setlength{\tabcolsep}{1.4mm}
\renewcommand{\arraystretch}{1.0}
\begin{tabular}{@{}l*{10}{c}c@{}}
\toprule
 & \multicolumn{5}{c}{\textit{Serialized handoff}} & \multicolumn{5}{c}{\textit{Parallel contention}} & \\
\cmidrule(lr){2-6}\cmidrule(lr){7-11}
\textbf{ID} & \textbf{Feat} & \textbf{Layer} & \textbf{Pipe} & \textbf{QA} & \textbf{Rot} & \textbf{PM} & \textbf{Swarm} & \textbf{OSS} & \textbf{Adv} & \textbf{Comp} & \textbf{Avg} \\
\midrule
P00 & 74.4 & 76.0 & 83.1 & 78.2 & 70.3 & 83.7 & 82.0 & 61.5 & 80.8 & 82.5 & \textbf{77.2} \\
P01 & 82.4 & 57.2 & 83.0 & 73.6 & 72.8 & 69.6 & 67.6 & 69.1 & 57.8 & 69.8 & \textbf{70.3} \\
P02 & 78.2 & 66.1 & 72.0 & 77.0 & 76.5 & 80.9 & 76.9 & 69.6 & 75.2 & 75.0 & \textbf{74.7} \\
P03 & 72.7 & 68.0 & 69.7 & 91.7 & 90.5 & 80.2 & 89.4 & 57.2 & 75.0 & 87.1 & \textbf{78.1} \\
P04 & 76.7 & 62.6 & 58.7 & 78.4 & 78.2 & 68.9 & 75.2 & 67.4 & 59.6 & 65.5 & \textbf{69.1} \\
P05 & 92.7 & 85.8 & 96.2 & 91.9 & 94.0 & 90.2 & 89.2 & 81.5 & 88.4 & 89.5 & \textbf{89.9} \\
P06 & 92.0 & 95.2 & 90.2 & 96.3 & 95.0 & 94.1 & 80.8 & 80.1 & 85.0 & 96.5 & \textbf{90.5} \\
P07 & 94.1 & 96.3 & 96.0 & 77.3 & 95.8 & 94.3 & 80.5 & 90.5 & 92.7 & 92.2 & \textbf{91.0} \\
P08 & 86.4 & 97.5 & 97.5 & 99.1 & 97.8 & 84.6 & 82.5 & 89.5 & 95.2 & 98.3 & \textbf{92.8} \\
P09 & 75.1 & 64.5 & 72.9 & 69.4 & 62.1 & 69.1 & 68.8 & 68.4 & 61.4 & 59.1 & \textbf{67.1} \\
\midrule
\textbf{Avg} & \textbf{82.5} & \textbf{76.9} & \textbf{81.9} & \textbf{83.3} & \textbf{83.3} & \textbf{81.6} & \textbf{79.3} & \textbf{73.5} & \textbf{77.1} & \textbf{81.5} & \textbf{80.1} \\
\bottomrule
\end{tabular}
\caption{Per-project $\times$ per-mode best score, averaged over the three
full-grid models (DeepSeek v4 Flash/Pro, GLM-5.2). Columns are grouped as
in the main-paper radar panels; the right column and bottom row give
project and mode means. The best mode per project is discussed in the
text.}
\label{tab:proj_mode}
\end{table*}

\noindent\textbf{No mode dominates; fit to the project is the effect.} Averaged over projects (Table~\ref{tab:proj_mode}), the modes span only 73.5 (open-source review) to 83.3 (QA-first and rotation tie), a range narrow enough that no single policy is globally best. Within a single project, however, the spread is far larger -- e.g. 57.2--91.7 on requirement tracking and 58.7--96.2 on e-commerce -- so coordination policy matters most when matched to the project's dependency structure. Serialized-handoff modes win projects with long requirement chains or clean stage boundaries, while parallel-contention modes pay a reconciliation tax that only competitive recovers, and only when redundant candidates can be cheaply scored.

\section{\textsc{LegoGent} Runtime: Bounded Synchronization}

\textsc{LegoGent} does not infer a fixed dependency graph. Instead it runs
a periodic sync loop over concurrently working agents, with two bounded
communication channels and a runtime-state completion gate.

\noindent\textbf{Passive sync channel.} Roughly every four minutes the
runtime broadcasts a consolidated \emph{progress-alignment snapshot} to
every agent. Each agent contributes four fields---completed work, test
evidence, blockers, and next steps---and receives the merged team view.
Figure~\ref{fig:loop} (top) is a translated snapshot from the
feature-squad / Claude Opus 4.8 / instant-messaging run (broadcast
interval $\approx$244\,s). Crucially, the snapshot is explicitly
\emph{not} a completion signal.

\noindent\textbf{Completion gate.} A round is accepted only when the
runtime state shows all agents \emph{idle and stable} \emph{and} the
required artifacts for the mode exist. This decouples ``the team stopped
talking'' from ``the team is done,'' and is what makes idle time and
merge repair observable rather than hidden.

\noindent\textbf{Active mailbox.} Between sync rounds, an agent may send a
targeted question or handoff directly to a peer. In this run
(Figure~\ref{fig:loop}), agents opened 52 such peer threads. This is the
channel that lets, e.g., a backend owner confirm a route shape with a
frontend owner without waiting for the next broadcast.

\begin{figure*}[t]
\begin{lstlisting}
=====  COLLABORATION: LegoGent periodic sync  ==============================================
[BTW_PROGRESS_ALIGNMENT #5]  periodic sync broadcast, ~every 244s; 52 peer-mailbox threads this run
NOTE: alignment snapshot, NOT a completion signal. A round enters user-review ONLY when the
runtime state is all-idle-and-stable AND the mode's required artifacts exist.

Runtime states:  dev-1: idle 244s   dev-2: idle 244s   dev-3: idle 244s   dev-4: idle 244s

Per-agent snapshot (1 done / 2 evidence / 3 blockers / 4 next):
 [dev-1] 1: all backend APIs + full frontend integrated on single port 8002.
         2: curl all-green; WebSocket realtime ~9ms; Playwright e2e 9/10 steps pass.
         3: dev-2/3/4 crashed on a provider API error; dev-1 took over the remaining frontend work.
         4: fix group-chat view activation; rerun Playwright; write docs (api-spec, modules, report).
 [dev-2] 1: conversation/message endpoints + WebSocket push done.
         2: curl branches (401/403/400/404) green; WS 8.9ms (<200ms).
         3: waiting on shell App.tsx before frontend build.
         4: mount ChatPage; dual-browser Playwright verify.
 [dev-3] 1: reply-count / single-message / history-filter / soft-delete endpoints landed, integrated by dev-1.
         2: selftest 36/36 assertions green incl. 401/403/404/422.
         3: not hard-blocked.
         4: build message-ops UI; coordinate mount point with dev-1.
 [dev-4] --: crashed on the provider API error; its remaining frontend work was absorbed by dev-1 (above).

=====  EVALUATION: TAgent grades the same round, then feeds a ranked repair agenda to round N+1  ====
# discovery_result.json -- reconstructed route surface (excerpt: method, path, auth, params)
POST   /api/users/register   auth:no   {username,password,nickname?}
POST   /api/users/login      auth:no   {username,password}
DELETE /api/users/me         auth:yes  {}            # cancel account
GET    /api/users/search     auth:yes  {keyword}
POST   /api/friends/request  auth:yes  {to_id,message?}

# API evaluator -- requirement: chat_send_text
scenario "send text"          weight 3.0  PASS
  user1 sent a text in conversation 4; user2 received immediately; unread count updated;
  retrievable via GET /api/conversations/4/messages with full body + metadata.
scenario "empty & long text"  weight 1.0  PASS
  empty message rejected 400 ("content must not be empty"); ~650-char accepted 200, body preserved.

# Code evaluator -- requirement: sys_security_no_plaintext
dir_filter: keep=[backend,docs,frontend,tests] drop=[.claude,.legogent]
scenario "HTTPS enforced"       weight 2.0  FAIL
  backend/main.py serves plain HTTP on port 8002; no TLS/redirect/HSTS.
scenario "sensitive transport"  weight 1.0  FAIL
  passwords sent as plaintext JSON (api.ts:32); bcrypt is server-side only
  (security.py:28) -> raw plaintext over unencrypted HTTP.
\end{lstlisting}
\caption{One \sys round end to end, from the feature-squad / Claude Opus
4.8 / instant-messaging run. \textbf{Top (collaboration):} a periodic
\textsc{LegoGent} sync broadcast---the runtime-state block and the
``idle-and-stable + required-artifacts'' gate are the actual basis for
accepting a round, and each agent reports the same four fields every
cycle; here dev-4 crashed on a provider API error and dev-1 absorbed its
frontend work, a real multi-agent failure the round still had to
reconcile. \textbf{Bottom (evaluation):} \textsc{TAgent} then grades the
deployed round---route discovery, a passing API check, and a failing
security check---with every verdict citing concrete behavior (status
codes, measured latency, \texttt{file:line}); this item-level evidence is
returned to \textsc{LegoGent} as the ranked repair agenda that opens the
next round.}
\label{fig:loop}
\end{figure*}

\section{\textsc{TAgent} Verification Methodology}

\textsc{TAgent} grades an unseen deployment by discovering its
implementation surface, routing each requirement to the appropriate
evaluator, and aggregating weighted evidence into a 100-point Functional
Completion Score.

\noindent\textbf{Step 1: Rubric compilation.} Each raw requirement
document is parsed into a structured rubric of weighted scenarios with
pass/partial/fail criteria. For P00 this yields 56 checks across three
categories (7 system-level, 46 functional, 3 documentation).

\noindent\textbf{Step 2: Surface discovery.} Given only the staged URL
and repository, \textsc{TAgent} reconstructs the route surface (from
OpenAPI when present, else from source). Each entry records method, path,
parameters, and whether authentication is required (the discovery block
of Figure~\ref{fig:loop}), so probing uses the real contract rather than
a guessed one.

\noindent\textbf{Step 3: Three evaluator classes.}
\begin{itemize}
\item \textbf{UI}: Playwright drives the live site,
reads the accessibility tree/DOM, and judges observed states. Stateful
flows (register\,$\to$\,login\,$\to$\,befriend\,$\to$\,enter room) are set
up once and reused. Destructive flows get isolated browser sessions.
\item \textbf{API}: authenticated HTTP sequences
against discovered routes; each check stores a request/response
\texttt{execution\_log} and a scored \texttt{checklist}.
\item \textbf{Code}: a directory filter
first keeps relevant folders and drops noise (e.g.\ \texttt{.claude},
\texttt{.legogent}), then targeted search and LLM inspection verify
constraints that are not externally visible, such as password hashing or
transport security.
\end{itemize}

\noindent\textbf{Step 4: Weighted checklists and evidence.} Every
requirement produces a checklist of weighted scenarios, each with a
verdict and a human-readable justification, plus raw evidence
(\texttt{api\_response}, \texttt{frontend\_test}, or \texttt{code\_snippet}).
The bottom of Figure~\ref{fig:loop} shows a passing API check and a
failing code check from the same run---note that verdicts cite concrete
behavior (measured latency, exact status codes, file:line).

\noindent\textbf{Step 5: Aggregation and feedback.} Scenario scores are
summed by rubric weight to the 100-point score. A requirement graded by
more than one evaluator (e.g.\ backend + frontend) is averaged per
\texttt{requirement\_id}, so a feature only counts as done when it holds
end to end. The same item-level evidence---observed state, failing
prerequisite, and affected weight---is returned to \textsc{LegoGent} as a
ranked repair agenda for the next round.

\section{Token and Cost Accounting}

The budget metric is computed from four token fields logged per agent and
per round: \emph{new-input}, \emph{cache-create}, \emph{cache-read}, and
\emph{output}. Cost is the price-weighted sum over these fields,
\[
\mathrm{USD} \;=\; \sum_{a,r} \big(
n^{\mathrm{in}}_{a,r}\,\pi_{\mathrm{in}}
+ n^{\mathrm{cc}}_{a,r}\,\pi_{\mathrm{cc}}
+ n^{\mathrm{cr}}_{a,r}\,\pi_{\mathrm{cr}}
+ n^{\mathrm{out}}_{a,r}\,\pi_{\mathrm{out}} \big),
\]
where the per-model rates $\pi_{\bullet}$ are the published provider list
prices in Table~\ref{tab:pricing}. Separating the four fields is what lets
\sys distinguish raw token volume from actual budget pressure: cache reads
are far cheaper than fresh input, so two runs with similar total tokens
can differ several-fold in dollars. Table~\ref{tab:tokens} gives the real
per-round breakdown for the worked example; the totals
($\approx$113\,M tokens, 110~min, best score 96.95) match its row in the
main results table. Table~\ref{tab:pricing} lists the four per-model
rates; applied to the token counts below they reproduce the USD cost
column of every results table---for instance the \$417.97 first-round
cost of this worked example.

\begin{table}[t]
\centering
\small
\setlength{\tabcolsep}{1.1mm}
\renewcommand{\arraystretch}{1.05}
\begin{tabular}{@{}lrrrr@{}}
\toprule
\textbf{Model} & \textbf{New-in} & \textbf{Cache-wr} & \textbf{Cache-rd} & \textbf{Out} \\
\midrule
Claude Opus 4.8   & 5.00 & 6.25 & 0.50   & 25.00 \\
GPT-5.5              & 5.00 & ---  & 0.50   & 30.00 \\
DeepSeek v4 Pro        & 0.45 & ---  & 0.0038 & 0.91 \\
DeepSeek v4 Flash     & 0.15 & ---  & 0.0030 & 0.30 \\
GLM-5.2                     & 1.21 & ---  & 0.30   & 4.24 \\
Qwen3.6-Flash                 & 0.18 & 0.23 & 0.018  & 1.09 \\
\bottomrule
\end{tabular}
\caption{Provider list prices used for the budget metric, in USD per
million tokens: new input, cache write, cache read, and output.
``---'' marks a field not billed as a separate line item (cache writes
fold into new input). Anthropic and OpenAI publish in USD; DeepSeek,
Z.AI (GLM-5.2), and Alibaba Cloud (Qwen) publish in CNY and are converted
at a fixed 6.6\,CNY/USD. Qwen3.6-Flash scored 0 across the grid and is
listed only for completeness.}
\label{tab:pricing}
\end{table}

\begin{table}[t]
\centering
\small
\setlength{\tabcolsep}{1.2mm}
\renewcommand{\arraystretch}{1.05}
\begin{tabular}{@{}lrrrrr@{}}
\toprule
\textbf{Round} & \textbf{New-in} & \textbf{Cache-cr} & \textbf{Cache-rd} & \textbf{Out} & \textbf{Total} \\
 & (k) & (M) & (M) & (k) & (M) \\
\midrule
R1 & 14.5 & 65.09 & 7.18 & 299.5 & 72.58 \\
R2 & \phantom{0}0.4 & 18.61 & 0.65 & \phantom{0}65.8 & 19.33 \\
R3 & \phantom{0}0.4 & 18.28 & 2.88 & \phantom{0}64.5 & 21.22 \\
\midrule
\textbf{Sum} & 15.3 & 101.98 & 10.71 & 429.8 & 113.13 \\
\bottomrule
\end{tabular}
\caption{Real per-round token accounting for feature squad / Claude Opus
4.8 / instant messaging (four agents). Cache-creation dominates, which is
why cache-aware pricing is required for a fair budget comparison.}
\label{tab:tokens}
\end{table}

\section{A Worked Refinement Trace}

Table~\ref{tab:traj} follows a subset of the 56 requirements across the
three rounds of the worked example, illustrating three phenomena
discussed in the main paper. First, mixed requirements are averaged across
evaluators (backend \texttt{be} / frontend \texttt{fe}), so a feature that
passes one side but not the other scores partially. Second, most R1
failures are repaired once \textsc{TAgent} feedback localizes them
(user-search, cancel-account, requirements-doc). Third, refinement is not
monotone: at R3 ``delete chat history'' regresses on the backend
(\texttt{be} $1\!\to\!0.5$) while other items are being fixed---the same
class of cross-requirement regression the main paper reports.

\begin{table}[t]
\centering
\small
\setlength{\tabcolsep}{1.1mm}
\renewcommand{\arraystretch}{1.05}
\begin{tabular}{@{}llccc@{}}
\toprule
\textbf{Requirement} & \textbf{Cat.} & \textbf{R1} & \textbf{R2} & \textbf{R3} \\
\midrule
Front--back communication & Sys & 10/10 & 10/10 & 10/10 \\
Message latency $<$200\,ms & Sys & 4/4 & 4/4 & 4/4 \\
Cancel account & Fun & 0/1 & 1/1 & 1/1 \\
Login/logout & Fun & 2.9/3 & 3/3 & 3/3 \\
User search & Fun & 1/2 & 2/2 & 2/2 \\
Delete chat history & Fun & 0.5/1 & 1/1 & \textbf{0.8/1} \\
No-plaintext transport & Sys & 0/4 & 0/4 & 2/4 \\
Requirements doc & Doc & 0/3 & 3/3 & 3/3 \\
\midrule
\textbf{Total} & & \textbf{79.22} & \textbf{93.07} & \textbf{96.95} \\
\bottomrule
\end{tabular}
\caption{Real per-requirement trajectory (excerpt) for the worked
example. Bold marks a backend regression at R3. The bottom row is the
aggregate Functional Completion Score.}
\label{tab:traj}
\end{table}

\section{Compute Environment}

All runs execute on four Intel(R) Xeon(R) Gold 6430 processors.
Deployment, testing, and evaluation use isolated port ranges so
concurrent configurations do not interfere. Each configuration allows up
to three refinement rounds; a round builds or patches the project,
deploys it through the \textsc{XDeploy} layer (which supplies TLS, HTTPS
redirects, and secure WebSockets), runs \textsc{TAgent}, and returns
scored feedback. Per-round wall-clock is capped (e.g.\ 90 minutes),
which is the source of some truncated first rounds analyzed in the main
paper.
\end{document}